\documentclass{article}
\usepackage{graphicx}
\usepackage{amsmath}
\usepackage{float}
\usepackage{booktabs}
\usepackage{multirow}
\usepackage{adjustbox}
\usepackage{url}
\usepackage[table]{xcolor}
\usepackage{makecell}

\graphicspath{{images/}}

\title{Machine Learnability as a Measure of Order in Aperiodic Sequences}

\author{Jennifer Dodgson, Michael Joedhitya, Adith Ramdas, \\Surender Suresh Kumar, Adarsh Singh Chauhan, \\ Akira Rafhael, Wang Mingshu and Nordine Lotfi* \\ }

\date{August 2025}

\begin{document}
\maketitle

\section{Abstract}
Research on the distribution of prime numbers has revealed a dual character: deterministic in definition yet exhibiting statistical behavior reminiscent of random processes. In this paper we show that it is possible to use an image-focused machine learning model to measure the comparative regularity of prime number fields at specific regions of an Ulam spiral. Specifically, we demonstrate that in pure accuracy terms, models trained on blocks extracted from regions of the spiral in the vicinity of 500m outperform models trained on blocks extracted from the region representing integers lower than 25m. This implies the existence of more easily learnable order in the former region than in the latter. Moreover, a detailed breakdown of precision and recall scores seem to imply that the model is favouring a different approach to classification in different regions of the spiral, focusing more on identifying prime patterns at lower numbers and more on eliminating composites at higher numbers. This aligns with number theory conjectures suggesting that at higher orders of magnitude we should see diminishing noise in prime number distributions, with averages (density, AP equidistribution) coming to dominate, while local randomness regularises after scaling by $\log x$. Taken together, these findings point toward an interesting possibility: that machine learning can serve as a new experimental instrument for number theory. Notably, the method shows potential for investigating the patterns in strong and weak primes for cryptographic purposes.

\section{Quasi-Order Among the Primes}
The Prime Number Theorem establishes the global density of primes as $\tfrac{x}{\log x}$ [1, 2], while average–equidistribution in arithmetic progressions is captured by the Bombieri–Vinogradov theorem [3, 4]; together these results formalize the “smooth” large-scale thinning of primes despite jagged local behavior. At the structural level, Green and Tao proved that the primes contain arbitrarily long arithmetic progressions [5], showing that strong additive structure persists at all scales. On the local side, Cramér’s probabilistic model predicts exponential-type statistics for normalized prime gaps and suggests maximal gaps of order $(\log x)^2$ [6], while the Hardy–Littlewood $k$-tuple conjecture supplies precise asymptotics for fixed prime constellations [7]. A deeper account of fluctuations connects primes to the spectrum of the Riemann zeta function: Montgomery’s pair-correlation work (assuming RH) [8] and Odlyzko’s large-scale computations [9] exhibit the same spacing statistics as random Hermitian matrices, indicating that fine-scale irregularities obey universal laws from random matrix theory. Related “randomness” phenomena for multiplicative functions—central to modeling quasi-randomness in prime patterns—are framed by the Chowla and Möbius-disjointness conjectures [10]; recent partial progress (e.g., logarithmically averaged results) supports the view that, after appropriate normalization, prime patterns converge toward stable statistical distributions.

\section{Hypothesis}
We hypothesize that while large-scale structural regularities in the distribution of prime numbers may only become fully apparent at magnitudes beyond polynomial-time computability, if the conjectured noise-reduction effects are continuous, traces of these regularities should be detectable at much smaller scales. In particular, machine learning models trained to learn dense representations of relational structure should exhibit a greater ability to learn of prime patterns at higher numerical ranges. Thus, if the conjectures hold, we expect models trained on larger primes to both achieve better performance than those trained at lower numbers, and to generalise better to out-of-distribution data taken from other regions of the sequence than those trained on smaller primes.

\section{Method}
To test this hypothesis, we generated black-and-white Ulam spiral images encoding consecutive integer sets:
\begin{itemize}
    \item \( n < 25{,}010{,}001 \) (5001×5001 pixels),
    \item \( 25{,}010{,}001 \leq n < 50{,}027{,}329 \) (7073×7073 pixels),
    \item \( 50{,}027{,}329 \leq n < 100{,}020{,}001 \) (10,001×10,001 pixels),
    \item \( 100{,}020{,}001 \leq n < 200{,}024{,}449 \) (14,143×14,143 pixels),
    \item \( 200{,}024{,}449 \leq n < 300{,}017{,}041 \) (17,321×17,321 pixels).
     \item \( 300{,}020{,}041 \leq n < 400{,}040{,}001 \) (20,001×20,001 pixels), and
    \item \( 400{,}040{,}001 \leq n < 500{,}014{,}321 \) (22,361×22,361 pixels).
\end{itemize}

Each image was subsampled into 350 non-overlapping blocks of size $256 \times 256$ pixels, $256 \times 256$ having been selected as a size favouring the learning of both local and global patterns. Blocks were divided into training (300) and validation (50) sets. Augmentation via rotation was tested but found to have either no or a negative result in most cases, and was thus abandoned in later trials.

\begin{figure}[H]
    \centering
    \includegraphics[width=1\linewidth]{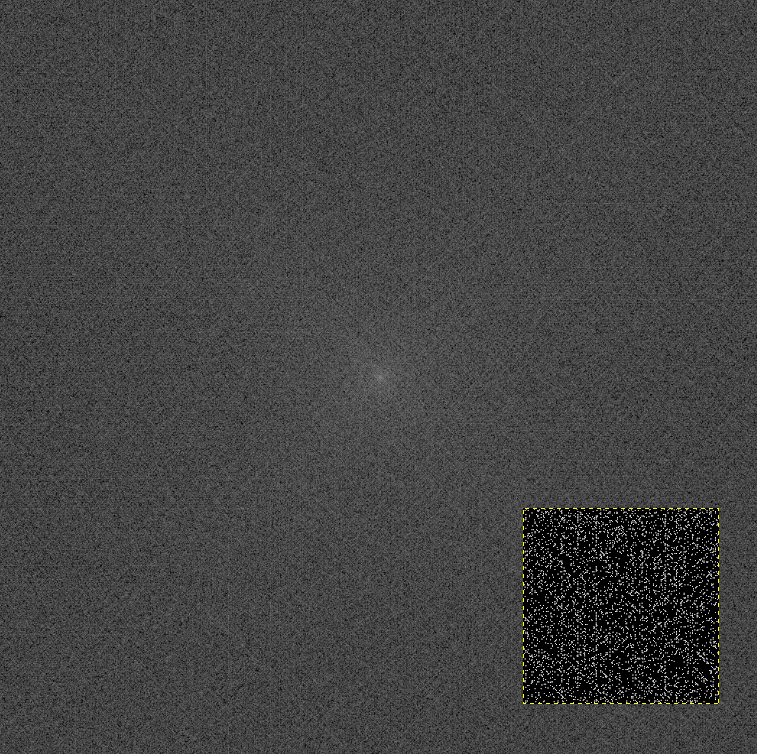}
    \caption{25,010,001-integer Ulam spiral, with $256 \times 256$ block (inset) for comparison.}
    \label{fig:ulam_spiral_25m}
\end{figure}

We trained a two-dimensional U-Net with a ResNet-34 encoder pre-initialized on ImageNet. This model architecture was chosen as a simple and effective implementation with the aim of favouring easy replication. While larger or more complex models will almost certainly produce higher overall accuracy scores, the goal of the present study was not necessarily to achieve the best possible accuracy overall, but rather to compare accuracy rates in different regions of the spiral. The network accepted one-channel grayscale inputs and produced a sigmoid-activated probability map, serving as a binary pixel-wise predictor for self-supervised inpainting from sparsified inputs. For each batch, we sampled an i.i.d. Bernoulli mask with mask ratio 0.3, revealing ~30\% of pixels and zeroing the remainder. The model was thus required to reconstruct the full binary image from a partially observed input.

The training objective was a weighted combination of soft mean class accuracy (Soft-MCA) and binary cross-entropy (BCE):

\[
\mathcal{L} = \alpha \,\mathcal{L}_{\text{Soft-MCA}} + \beta \,\text{BCE}, 
\quad (\alpha = 1.0,\ \beta = 0.5).
\]

Soft-MCA was computed using probabilistic outputs, averaging per-class accuracies to mitigate imbalance, while BCE was evaluated against binary ground truth. For evaluation, we additionally reported hard-thresholded (0.5) mean class accuracy.

Optimization was performed with Adam (learning rate $1 \times 10^{-4}$), batch size 4, for up to 150 epochs, with ReduceLROnPlateau (mode=‘min’, factor=0.5) and scheduling on validation loss.

An identical model was trained on each numerical range and subsequently evaluated across all test sets. Comparative performance was used as a proxy for the learnability of prime distributions within each numerical range.

The dataset generation/model training process was repeated three times to confirm broad trends and an average taken to minimise chaotic fluctuations in output at the individual model level. The results given below are thus the average of three identical models trained on different randomly shuffled subsets of the same dataset.

\section{Results}
Due to the substantial class imbalance in the dataset (with white/prime pixels comprising $\approx 6\%$ of low-integer blocks and $\approx 5\%$ of high-integer blocks), raw accuracy is not necessarily an informative performance measure. In particular, a trivial baseline that predicts all pixels as black achieves $\approx 94$--$96\%$ accuracy. We therefore report precision, recall, F1 score, and confusion metrics with respect to the minority (prime) class to provide a more detailed breakdown of how results were produced.

For each evaluation, the model produced five outputs: (i) the masked input, (ii) the inpainting reconstruction, (iii) the pixel-wise error map, (iv) the ground-truth image, and (v) summary statistics. 

\begin{figure}[H]
    \centering
    \includegraphics[width=1\linewidth]{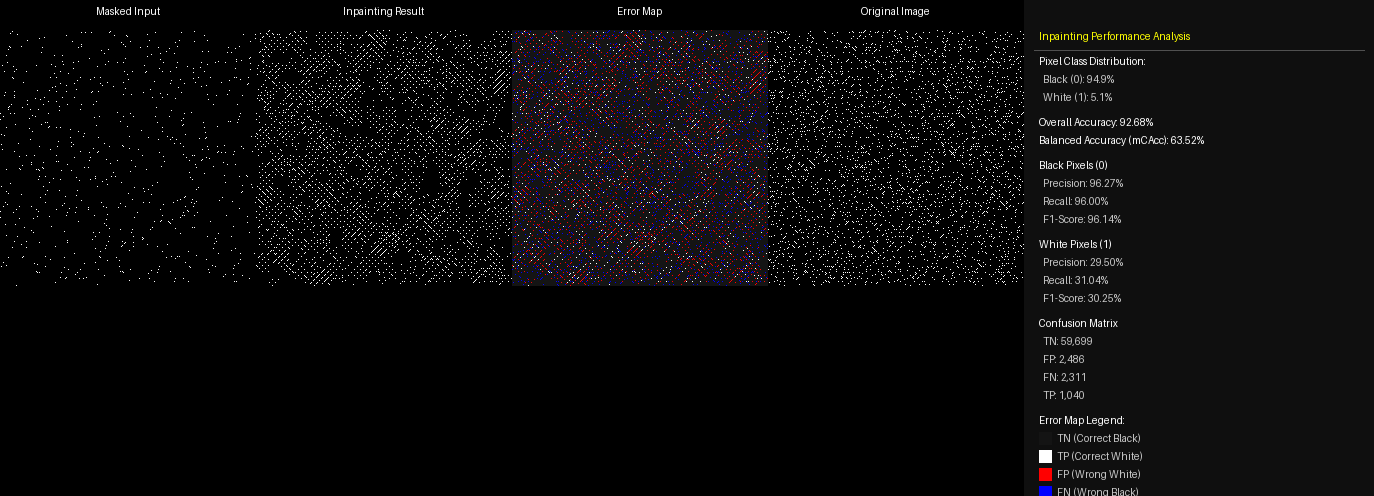}
    \caption{Example masked input, inpainting result, error map, original image and relevant statistics from a model trained around 100,000,000 and tested around 300,000,000.}
    \label{fig:example_output}
\end{figure}

To account for the limited dataset size, we employed resampling with $n = 10{,}000$ replicates and computed nonparametric block-level bootstrap confidence intervals for all reported metrics, enabling a more reliable assessment of statistical significance.

\begin{table}[H]
\centering
\begin{adjustbox}{width=\textwidth,center}
\begin{tabular}{c|ccccccc}
\toprule
\multirow{2}{*}{\textbf{Test set}} & \multicolumn{5}{c}{\textbf{Train set}} \\
\cmidrule(lr){2-8}
 & 25m & 50m & 100m & 200m & 300m & 400m & 500m \\
\midrule
25m  & 
\cellcolor[HTML]{bfcb74} 0.8768 (±0.0029) & 
\cellcolor[HTML]{c7cd72} 0.8788 (±0.0019) & 
\cellcolor[HTML]{d8cf6f} 0.8826 (±0.0016) & 
\cellcolor[HTML]{e1d16d} 0.8846 (±0.0014) & 
\cellcolor[HTML]{e1d16d} 0.8846 (±0.0016) & 
\cellcolor[HTML]{e5d16c} 0.8856 (±0.0016) & 
\cellcolor[HTML]{f0d36a} 0.8881 (±0.0014) \\
50m  & 
\cellcolor[HTML]{b7ca76} 0.8749 (±0.0049) & 
\cellcolor[HTML]{dfd16d} 0.8843 (±0.0039) & 
\cellcolor[HTML]{fdcf67} 0.8926 (±0.0030) & 
\cellcolor[HTML]{f7b76a} 0.8965 (±0.0028) & 
\cellcolor[HTML]{f6b36b} 0.8973 (±0.0029) & 
\cellcolor[HTML]{f2a86c} 0.8990 (±0.0025) & 
\cellcolor[HTML]{ee986f} 0.9016 (±0.0016) \\
100m & 
\cellcolor[HTML]{92c47e} 0.8666 (±0.0072) & 
\cellcolor[HTML]{cece71} 0.8802 (±0.0054) & 
\cellcolor[HTML]{fdcd67} 0.8930 (±0.0036) & 
\cellcolor[HTML]{f7b96a} 0.8962 (±0.0043) & 
\cellcolor[HTML]{f5b16b} 0.8975 (±0.0041) & 
\cellcolor[HTML]{f0a06d} 0.9003 (±0.0033) & 
\cellcolor[HTML]{eb8b70} 0.9037 (±0.0017) \\
200m & 
\cellcolor[HTML]{57bb8a} 0.8529 (±0.0080) & 
\cellcolor[HTML]{9ec67b} 0.8692 (±0.0067) & 
\cellcolor[HTML]{e3d16c} 0.8851 (±0.0050) & 
\cellcolor[HTML]{fbd567} 0.8907 (±0.0048) & 
\cellcolor[HTML]{fed567} 0.8914 (±0.0058) & 
\cellcolor[HTML]{f9c169} 0.8949 (±0.0050) & 
\cellcolor[HTML]{f3ab6c} 0.8985 (±0.0031) \\
300m & 
\cellcolor[HTML]{9bc57c} 0.8686 (±0.0057) & 
\cellcolor[HTML]{cdce71} 0.8801 (±0.0057) & 
\cellcolor[HTML]{fed166} 0.8923 (±0.0047) & 
\cellcolor[HTML]{f1a36d} 0.8998 (±0.0033) & 
\cellcolor[HTML]{f2a56d} 0.8995 (±0.0044) & 
\cellcolor[HTML]{eb8d70} 0.9033 (±0.0032) & 
\cellcolor[HTML]{e67c73} 0.9061 (±0.0025) \\
400m & 
\cellcolor[HTML]{64bd88} 0.8560 (±0.0060) & 
\cellcolor[HTML]{a5c77a} 0.8710 (±0.0062) & 
\cellcolor[HTML]{e1d16d} 0.8846 (±0.0052) & 
\cellcolor[HTML]{ffd666} 0.8914 (±0.0056) & 
\cellcolor[HTML]{fac468} 0.8944 (±0.0050) & 
\cellcolor[HTML]{f4ad6c} 0.8982 (±0.0047) & 
\cellcolor[HTML]{ee976f} 0.9017 (±0.0033) \\
500m & 
\cellcolor[HTML]{63bd88} 0.8559 (±0.0054) & 
\cellcolor[HTML]{a9c879} 0.8717 (±0.0057) & 
\cellcolor[HTML]{e7d26b} 0.8861 (±0.0053) & 
\cellcolor[HTML]{fbc868} 0.8938 (±0.0051) & 
\cellcolor[HTML]{f6b36b} 0.8971 (±0.0051) & 
\cellcolor[HTML]{f1a46d} 0.8996 (±0.0045) & 
\cellcolor[HTML]{eb8c70} 0.9035 (±0.0038) \\
\bottomrule
\end{tabular}
\end{adjustbox}
\caption{Overall accuracy/micro F1. Averaged performance across three models.}
\label{tab:overall_accuracy_micro_f1}
\end{table}

\begin{table}[H]
\centering
\begin{adjustbox}{width=\textwidth,center}
\begin{tabular}{c|ccccccc}
\toprule
\multirow{2}{*}{\textbf{Test set}} & \multicolumn{5}{c}{\textbf{Train set}} \\
\cmidrule(lr){2-8}
& 25m & 50m & 100m & 200m & 300m & 400m & 500m \\
\midrule
25m & 
\cellcolor[HTML]{e67c73} 0.6246 ($\pm$0.0039) & 
\cellcolor[HTML]{f1a36d} 0.6126 ($\pm$0.0031) & 
\cellcolor[HTML]{f7b76a} 0.6062 ($\pm$0.0025) & 
\cellcolor[HTML]{fac268} 0.6026 ($\pm$0.0025) & 
\cellcolor[HTML]{fbc668} 0.6016 ($\pm$0.0030) & 
\cellcolor[HTML]{fccb67} 0.5999 ($\pm$0.0030) & 
\cellcolor[HTML]{ffd466} 0.5971 ($\pm$0.0020) \\
50m & 
\cellcolor[HTML]{f1a26d} 0.6127 ($\pm$0.0038) & 
\cellcolor[HTML]{f7b76a} 0.6063 ($\pm$0.0039) & 
\cellcolor[HTML]{fed066} 0.5983 ($\pm$0.0033) & 
\cellcolor[HTML]{e0d16d} 0.5937 ($\pm$0.0031) & 
\cellcolor[HTML]{d3cf70} 0.5926 ($\pm$0.0034) & 
\cellcolor[HTML]{bbcb75} 0.5906 ($\pm$0.0030) & 
\cellcolor[HTML]{9bc67c} 0.5879 ($\pm$0.0019) \\
100m & 
\cellcolor[HTML]{f1a26d} 0.6129 ($\pm$0.0032) & 
\cellcolor[HTML]{f7b66a} 0.6065 ($\pm$0.0039) & 
\cellcolor[HTML]{fed166} 0.5982 ($\pm$0.0036) & 
\cellcolor[HTML]{d9cf6f} 0.5931 ($\pm$0.0036) & 
\cellcolor[HTML]{cece71} 0.5922 ($\pm$0.0039) & 
\cellcolor[HTML]{b6ca76} 0.5901 ($\pm$0.0038) & 
\cellcolor[HTML]{8bc37f} 0.5865 ($\pm$0.0020) \\
200m & 
\cellcolor[HTML]{f2a66d} 0.6117 ($\pm$0.0028) & 
\cellcolor[HTML]{f6b66a} 0.6067 ($\pm$0.0029) & 
\cellcolor[HTML]{fdcd67} 0.5993 ($\pm$0.0031) & 
\cellcolor[HTML]{ffd666} 0.5963 ($\pm$0.0036) & 
\cellcolor[HTML]{eed36a} 0.5950 ($\pm$0.0037) & 
\cellcolor[HTML]{d8cf6f} 0.5931 ($\pm$0.0034) & 
\cellcolor[HTML]{b2c977} 0.5898 ($\pm$0.0028) \\
300m & 
\cellcolor[HTML]{f3a86c} 0.6109 ($\pm$0.0031) & 
\cellcolor[HTML]{f9be69} 0.6040 ($\pm$0.0035) & 
\cellcolor[HTML]{f5d469} 0.5955 ($\pm$0.0043) & 
\cellcolor[HTML]{abc878} 0.5892 ($\pm$0.0035) & 
\cellcolor[HTML]{a8c879} 0.5890 ($\pm$0.0045) & 
\cellcolor[HTML]{87c280} 0.5862 ($\pm$0.0035) & 
\cellcolor[HTML]{57bb8a} 0.5820 ($\pm$0.0025) \\
400m & 
\cellcolor[HTML]{f3a96c} 0.6108 ($\pm$0.0029) & 
\cellcolor[HTML]{f7b96a} 0.6056 ($\pm$0.0033) & 
\cellcolor[HTML]{ffd666} 0.5964 ($\pm$0.0038) & 
\cellcolor[HTML]{cfce71} 0.5923 ($\pm$0.0044) & 
\cellcolor[HTML]{b4c977} 0.5900 ($\pm$0.0044) & 
\cellcolor[HTML]{a0c67b} 0.5883 ($\pm$0.0043) & 
\cellcolor[HTML]{6bbe86} 0.5838 ($\pm$0.0027) \\
500m & 
\cellcolor[HTML]{f3aa6c} 0.6104 ($\pm$0.0025) & 
\cellcolor[HTML]{f8ba6a} 0.6052 ($\pm$0.0028) & 
\cellcolor[HTML]{ffd366} 0.5975 ($\pm$0.0033) & 
\cellcolor[HTML]{cece71} 0.5922 ($\pm$0.0036) & 
\cellcolor[HTML]{bccb75} 0.5907 ($\pm$0.0038) & 
\cellcolor[HTML]{a2c77a} 0.5884 ($\pm$0.0037) & 
\cellcolor[HTML]{7dc182} 0.5853 ($\pm$0.0032) \\
\bottomrule
\end{tabular}
\end{adjustbox}
\caption{Macro F1. Averaged performance across three models.}
\label{tab:macro_f1}
\end{table}

\begin{table}[H]
\centering
\begin{adjustbox}{width=\textwidth,center}
\begin{tabular}{c|ccccccc}
\toprule
\multirow{2}{*}{\textbf{Test set}} & \multicolumn{5}{c}{\textbf{Train set}} \\
\cmidrule(lr){2-8}
 & 25m & 50m & 100m & 200m & 300m & 400m & 500m \\
\midrule
25m & 
\cellcolor[HTML]{e67c73} 0.2429 ($\pm$0.0051) & 
\cellcolor[HTML]{f3a96c} 0.2240 ($\pm$0.0046) & 
\cellcolor[HTML]{fac268} 0.2131 ($\pm$0.0040) & 
\cellcolor[HTML]{fed166} 0.2065 ($\pm$0.0044) & 
\cellcolor[HTML]{f1d369} 0.2034 ($\pm$0.0046) & 
\cellcolor[HTML]{cbcd72} 0.2007 ($\pm$0.0046) & 
\cellcolor[HTML]{85c281} 0.1958 ($\pm$0.0032) \\
50m & 
\cellcolor[HTML]{f1a26d} 0.2269 ($\pm$0.0044) & 
\cellcolor[HTML]{f5b16b} 0.2203 ($\pm$0.0044) & 
\cellcolor[HTML]{fbc868} 0.2107 ($\pm$0.0042) & 
\cellcolor[HTML]{ffd666} 0.2044 ($\pm$0.0041) & 
\cellcolor[HTML]{d7cf6f} 0.2016 ($\pm$0.0042) & 
\cellcolor[HTML]{c0cb74} 0.2000 ($\pm$0.0043) & 
\cellcolor[HTML]{79c083} 0.1950 ($\pm$0.0035) \\
100m & 
\cellcolor[HTML]{f3aa6c} 0.2236 ($\pm$0.0038) & 
\cellcolor[HTML]{f7b76a} 0.2177 ($\pm$0.0038) & 
\cellcolor[HTML]{fbc768} 0.2111 ($\pm$0.0040) & 
\cellcolor[HTML]{ffd666} 0.2044 ($\pm$0.0035) & 
\cellcolor[HTML]{d5cf6f} 0.2014 ($\pm$0.0045) & 
\cellcolor[HTML]{cccd71} 0.2009 ($\pm$0.0041) & 
\cellcolor[HTML]{7fc182} 0.1954 ($\pm$0.0031) \\
200m & 
\cellcolor[HTML]{f6b56a} 0.2188 ($\pm$0.0036) &
\cellcolor[HTML]{fac268} 0.2131 ($\pm$0.0033) & 
\cellcolor[HTML]{fed266} 0.2062 ($\pm$0.0038) & 
\cellcolor[HTML]{e8d26b} 0.2028 ($\pm$0.0036) & 
\cellcolor[HTML]{b4ca76} 0.1991 ($\pm$0.0035) & 
\cellcolor[HTML]{afc978} 0.1988 ($\pm$0.0035) & 
\cellcolor[HTML]{68bd87} 0.1938 ($\pm$0.0030) \\
300m & 
\cellcolor[HTML]{f4ac6c} 0.2227 ($\pm$0.0041) & 
\cellcolor[HTML]{f9be69} 0.2148 ($\pm$0.0038) &
\cellcolor[HTML]{fdce67} 0.2080 ($\pm$0.0037) & 
\cellcolor[HTML]{ffd666} 0.2044 ($\pm$0.0042) & 
\cellcolor[HTML]{c8cd72} 0.2006 ($\pm$0.0044) & 
\cellcolor[HTML]{c5cc73} 0.2003 ($\pm$0.0045) & 
\cellcolor[HTML]{70bf85} 0.1943 ($\pm$0.0031) \\
400m & 
\cellcolor[HTML]{f7b86a} 0.2174 ($\pm$0.0038) & 
\cellcolor[HTML]{fac468} 0.2121 ($\pm$0.0035) & 
\cellcolor[HTML]{ffd566} 0.2052 ($\pm$0.0038) & 
\cellcolor[HTML]{cacd72} 0.2007 ($\pm$0.0034) & 
\cellcolor[HTML]{a7c779} 0.1982 ($\pm$0.0043) & 
\cellcolor[HTML]{a0c67b} 0.1977 ($\pm$0.0040) & 
\cellcolor[HTML]{57bb8a} 0.1926 ($\pm$0.0032) \\
500m & 
\cellcolor[HTML]{f8ba6a} 0.2167 ($\pm$0.0040) & 
\cellcolor[HTML]{fbc768} 0.2111 ($\pm$0.0038) & 
\cellcolor[HTML]{ffd666} 0.2046 ($\pm$0.0037) & 
\cellcolor[HTML]{c3cc73} 0.2002 ($\pm$0.0037) & 
\cellcolor[HTML]{afc978} 0.1988 ($\pm$0.0036) & 
\cellcolor[HTML]{99c57c} 0.1972 ($\pm$0.0039) & 
\cellcolor[HTML]{5fbc89} 0.1931 ($\pm$0.0031) \\
\bottomrule
\end{tabular}
\end{adjustbox}
\caption{White (prime) precision. Averaged performance across three models.}
\label{tab:white_prime_precision}
\end{table}

\begin{table}[H]
\centering
\begin{adjustbox}{width=\textwidth,center}
\begin{tabular}{c|ccccccc}
\toprule
\multirow{2}{*}{\textbf{Test set}} & \multicolumn{5}{c}{\textbf{Train set}} \\
\cmidrule(lr){2-8}
& 25m & 50m & 100m & 200m & 300m & 400m & 500m \\
\midrule
25m & 
\cellcolor[HTML]{ee996e} 0.4571 ($\pm$0.0209) & 
\cellcolor[HTML]{f4ac6c} 0.4166 ($\pm$0.0143) & 
\cellcolor[HTML]{f6b66a} 0.3945 ($\pm$0.0118) & 
\cellcolor[HTML]{f8bb69} 0.3828 ($\pm$0.0103) & 
\cellcolor[HTML]{f8bb6a} 0.3844 ($\pm$0.0115) & 
\cellcolor[HTML]{f9be69} 0.3759 ($\pm$0.0123) & 
\cellcolor[HTML]{fac468} 0.3644 ($\pm$0.0099) \\
50m & 
\cellcolor[HTML]{f4ac6c} 0.4162 ($\pm$0.0282) & 
\cellcolor[HTML]{fac269} 0.3690 ($\pm$0.0246) & 
\cellcolor[HTML]{ffd666} 0.3235 ($\pm$0.0178) & 
\cellcolor[HTML]{d2ce70} 0.3038 ($\pm$0.0172) & 
\cellcolor[HTML]{c5cc73} 0.3010 ($\pm$0.0166) & 
\cellcolor[HTML]{a5c77a} 0.2910 ($\pm$0.0142) & 
\cellcolor[HTML]{7ec182} 0.2801 ($\pm$0.0088) \\
100m & 
\cellcolor[HTML]{ef996e} 0.4567 ($\pm$0.0383) & 
\cellcolor[HTML]{f7b96a} 0.3872 ($\pm$0.0310) & 
\cellcolor[HTML]{fad568} 0.3213 ($\pm$0.0212) & 
\cellcolor[HTML]{cece71} 0.3018 ($\pm$0.0237) & 
\cellcolor[HTML]{c5cc73} 0.2980 ($\pm$0.0239) & 
\cellcolor[HTML]{a5c77a} 0.2835 ($\pm$0.0199) & 
\cellcolor[HTML]{7ec182} 0.2663 ($\pm$0.0101) \\
200m & 
\cellcolor[HTML]{e67c73} 0.5200 ($\pm$0.0405) & 
\cellcolor[HTML]{f1a26d} 0.4387 ($\pm$0.0357) & 
\cellcolor[HTML]{fbc768} 0.3578 ($\pm$0.0269) & 
\cellcolor[HTML]{fed266} 0.3334 ($\pm$0.0276) & 
\cellcolor[HTML]{ffd466} 0.3298 ($\pm$0.0316) & 
\cellcolor[HTML]{e7d26b} 0.3132 ($\pm$0.0266) & 
\cellcolor[HTML]{c0cb74} 0.2955 ($\pm$0.0171) \\
300m & 
\cellcolor[HTML]{f1a16d} 0.4412 ($\pm$0.0317) & 
\cellcolor[HTML]{f9bd69} 0.3781 ($\pm$0.0308) & 
\cellcolor[HTML]{f0d36a} 0.3171 ($\pm$0.0271) & 
\cellcolor[HTML]{9cc67c} 0.2797 ($\pm$0.0192) & 
\cellcolor[HTML]{9fc67b} 0.2810 ($\pm$0.0259) & 
\cellcolor[HTML]{79c083} 0.2639 ($\pm$0.0179) & 
\cellcolor[HTML]{57bb8a} 0.2487 ($\pm$0.0128) \\
400m & 
\cellcolor[HTML]{e98671} 0.4994 ($\pm$0.0315) & 
\cellcolor[HTML]{f3a86c} 0.4252 ($\pm$0.0318) & 
\cellcolor[HTML]{fcc967} 0.3529 ($\pm$0.0279) & 
\cellcolor[HTML]{f7d468} 0.3202 ($\pm$0.0308) & 
\cellcolor[HTML]{d9cf6f} 0.3067 ($\pm$0.0280) & 
\cellcolor[HTML]{afc978} 0.2881 ($\pm$0.0258) & 
\cellcolor[HTML]{88c380} 0.2710 ($\pm$0.0170) \\
500m & 
\cellcolor[HTML]{e98771} 0.4977 ($\pm$0.0266) & 
\cellcolor[HTML]{f3ab6c} 0.4194 ($\pm$0.0299) & 
\cellcolor[HTML]{fdcc67} 0.3465 ($\pm$0.0291) & 
\cellcolor[HTML]{dbd06e} 0.3077 ($\pm$0.0271) & 
\cellcolor[HTML]{bdcb75} 0.2943 ($\pm$0.0273) & 
\cellcolor[HTML]{9dc67b} 0.2799 ($\pm$0.0241) & 
\cellcolor[HTML]{75bf84} 0.2625 ($\pm$0.0203) \\
\bottomrule
\end{tabular}
\end{adjustbox}
\caption{White (prime) recall. Averaged performance across three models.}
\label{tab:white_prime_recall}
\end{table}

\begin{table}[H]
\centering
\begin{adjustbox}{width=\textwidth,center}
\begin{tabular}{c|ccccccc}
\toprule
\multirow{2}{*}{\textbf{Test set}} & \multicolumn{5}{c}{\textbf{Train set}} \\
\cmidrule(lr){2-8}
& 25m & 50m & 100m & 200m & 300m & 400m & 500m \\
\midrule
25m & 
\cellcolor[HTML]{e67c73} 0.3169 ($\pm$0.0090) & 
\cellcolor[HTML]{f0a06d} 0.2914 ($\pm$0.0071) & 
\cellcolor[HTML]{f6b56a} 0.2763 ($\pm$0.0060) & 
\cellcolor[HTML]{f9c169} 0.2677 ($\pm$0.0058) & 
\cellcolor[HTML]{fac368} 0.2657 ($\pm$0.0066) & 
\cellcolor[HTML]{fcc967} 0.2613 ($\pm$0.0066) & 
\cellcolor[HTML]{ffd366} 0.2543 ($\pm$0.0045) \\
50m & 
\cellcolor[HTML]{ef9c6e} 0.2939 ($\pm$0.0099) & 
\cellcolor[HTML]{f6b66a} 0.2755 ($\pm$0.0102) & 
\cellcolor[HTML]{ffd366} 0.2543 ($\pm$0.0085) & 
\cellcolor[HTML]{d7cf6f} 0.2430 ($\pm$0.0079) & 
\cellcolor[HTML]{cccd71} 0.2405 ($\pm$0.0082) & 
\cellcolor[HTML]{b5ca76} 0.2354 ($\pm$0.0072) & 
\cellcolor[HTML]{95c57d} 0.2283 ($\pm$0.0049) \\
100m & 
\cellcolor[HTML]{ed956f} 0.2994 ($\pm$0.0102) & 
\cellcolor[HTML]{f5b26b} 0.2784 ($\pm$0.0106) & 
\cellcolor[HTML]{ffd466} 0.2539 ($\pm$0.0092) & 
\cellcolor[HTML]{d2ce70} 0.2418 ($\pm$0.0093) & 
\cellcolor[HTML]{c7cd72} 0.2394 ($\pm$0.0102) & 
\cellcolor[HTML]{adc878} 0.2337 ($\pm$0.0093) & 
\cellcolor[HTML]{83c281} 0.2241 ($\pm$0.0052) \\
200m & 
\cellcolor[HTML]{eb8c70} 0.3058 ($\pm$0.0091) & 
\cellcolor[HTML]{f3a86c} 0.2856 ($\pm$0.0098) & 
\cellcolor[HTML]{fcca67} 0.2611 ($\pm$0.0090) & 
\cellcolor[HTML]{ffd666} 0.2518 ($\pm$0.0099) & 
\cellcolor[HTML]{eed36a} 0.2482 ($\pm$0.0108) & 
\cellcolor[HTML]{d6cf6f} 0.2427 ($\pm$0.0095) & 
\cellcolor[HTML]{afc978} 0.2339 ($\pm$0.0071) \\
300m & 
\cellcolor[HTML]{ef9b6e} 0.2949 ($\pm$0.0097) & 
\cellcolor[HTML]{f7b96a} 0.2735 ($\pm$0.0104) & 
\cellcolor[HTML]{f2d369} 0.2490 ($\pm$0.0115) & 
\cellcolor[HTML]{a6c779} 0.2321 ($\pm$0.0090) &
\cellcolor[HTML]{a3c77a} 0.2314 ($\pm$0.0118) & 
\cellcolor[HTML]{82c281} 0.2240 ($\pm$0.0087) & 
\cellcolor[HTML]{57bb8a} 0.2142 ($\pm$0.0062) \\
400m& 
\cellcolor[HTML]{ec916f} 0.3018 ($\pm$0.0089) & 
\cellcolor[HTML]{f4ad6b} 0.2817 ($\pm$0.0094) & 
\cellcolor[HTML]{fed166} 0.2555 ($\pm$0.0099) & 
\cellcolor[HTML]{d7cf6f} 0.2431 ($\pm$0.0120) & 
\cellcolor[HTML]{bbcb75} 0.2367 ($\pm$0.0113) & 
\cellcolor[HTML]{a2c77a} 0.2311 ($\pm$0.0112) & 
\cellcolor[HTML]{72bf85} 0.2204 ($\pm$0.0072) \\
500m & 
\cellcolor[HTML]{ed926f} 0.3012 ($\pm$0.0071) & 
\cellcolor[HTML]{f4af6b} 0.2805 ($\pm$0.0087) &
\cellcolor[HTML]{fed066} 0.2567 ($\pm$0.0098) & 
\cellcolor[HTML]{d0ce71} 0.2414 ($\pm$0.0100) & 
\cellcolor[HTML]{bbcb75} 0.2368 ($\pm$0.0105) & 
\cellcolor[HTML]{9fc67b} 0.2305 ($\pm$0.0098) & 
\cellcolor[HTML]{78c083} 0.2218 ($\pm$0.0083) \\
\bottomrule
\end{tabular}
\end{adjustbox}
\caption{White (prime) F1. Averaged performance across three models.}
\label{tab:white_prime_f1}
\end{table}

\begin{table}[H]
\centering
\begin{adjustbox}{width=\textwidth,center}
\begin{tabular}{c|ccccccc}
\toprule
\multirow{2}{*}{\textbf{Test set}} & \multicolumn{5}{c}{\textbf{Train set}} \\
\cmidrule(lr){2-8}
& 25m & 50m & 100m & 200m & 300m & 400m & 500m \\
\midrule
25m & 
\cellcolor[HTML]{f5b16b} 0.9615 ($\pm$0.0008) & 
\cellcolor[HTML]{f8bc69} 0.9608 ($\pm$0.0007) & 
\cellcolor[HTML]{f4ad6c} 0.9618 ($\pm$0.0004) & 
\cellcolor[HTML]{f2a46d} 0.9623 ($\pm$0.0007) & 
\cellcolor[HTML]{f09d6e} 0.9627 ($\pm$0.0004) & 
\cellcolor[HTML]{f09d6e} 0.9627 ($\pm$0.0008) & 
\cellcolor[HTML]{ee976f} 0.9631 ($\pm$0.0007) \\
50m & 
\cellcolor[HTML]{e1d16d} 0.9587 ($\pm$0.0010) & 
\cellcolor[HTML]{b6ca76} 0.9582 ($\pm$0.0010) & 
\cellcolor[HTML]{98c57c} 0.9579 ($\pm$0.0007) & 
\cellcolor[HTML]{c1cc74} 0.9583 ($\pm$0.0007) & 
\cellcolor[HTML]{dbd06e} 0.9587 ($\pm$0.0004) & 
\cellcolor[HTML]{beb764} 0.9586 ($\pm$0.0003) & 
\cellcolor[HTML]{ffd666} 0.9591 ($\pm$0.0003) \\
100m & 
\cellcolor[HTML]{f7b96a} 0.9610 ($\pm$0.0016) & 
\cellcolor[HTML]{e1bd5c} 0.9592 ($\pm$0.0011) & 
\cellcolor[HTML]{a1c77a} 0.9580 ($\pm$0.0008) & 
\cellcolor[HTML]{b3c977} 0.9582 ($\pm$0.0008) & 
\cellcolor[HTML]{d6cf6f} 0.9586 ($\pm$0.0008) & 
\cellcolor[HTML]{aac879} 0.9581 ($\pm$0.0007) & 
\cellcolor[HTML]{d0ce70} 0.9585 ($\pm$0.0003) \\
200m & 
\cellcolor[HTML]{e67c73} 0.9648 ($\pm$0.0020) & 
\cellcolor[HTML]{d79962} 0.9618 ($\pm$0.0013) & 
\cellcolor[HTML]{fed166} 0.9594 ($\pm$0.0010) & 
\cellcolor[HTML]{fdcd67} 0.9597 ($\pm$0.0009) & 
\cellcolor[HTML]{fbc868} 0.9600 ($\pm$0.0010) & 
\cellcolor[HTML]{fdcd67} 0.9597 ($\pm$0.0008) & 
\cellcolor[HTML]{fccb67} 0.9598 ($\pm$0.0008) \\
300m & 
\cellcolor[HTML]{fbc568} 0.9602 ($\pm$0.0012) & 
\cellcolor[HTML]{c1cc74} 0.9584 ($\pm$0.0011) & 
\cellcolor[HTML]{7bc083} 0.9575 ($\pm$0.0009) & 
\cellcolor[HTML]{57bb8a} 0.9571 ($\pm$0.0006) & 
\cellcolor[HTML]{84c281} 0.9576 ($\pm$0.0009) & 
\cellcolor[HTML]{74bf84} 0.9574 ($\pm$0.0007) & 
\cellcolor[HTML]{95c57d} 0.9578 ($\pm$0.0007) \\
400m & 
\cellcolor[HTML]{ec8f70} 0.9636 ($\pm$0.0015) & 
\cellcolor[HTML]{f7b96a} 0.9610 ($\pm$0.0013) & 
\cellcolor[HTML]{fed066} 0.9595 ($\pm$0.0010) & 
\cellcolor[HTML]{ffd666} 0.9591 ($\pm$0.0010) & 
\cellcolor[HTML]{e7d26c} 0.9588 ($\pm$0.0009) & 
\cellcolor[HTML]{b7ca76} 0.9582 ($\pm$0.0008) & 
\cellcolor[HTML]{d6cf6f} 0.9586 ($\pm$0.0007) \\
500m & 
\cellcolor[HTML]{ed936f} 0.9634 ($\pm$0.0013) & 
\cellcolor[HTML]{f9c169} 0.9605 ($\pm$0.0014) & 
\cellcolor[HTML]{ffd666} 0.9591 ($\pm$0.0009) & 
\cellcolor[HTML]{b1c977} 0.9582 ($\pm$0.0008) & 
\cellcolor[HTML]{c0cb74} 0.9583 ($\pm$0.0008) & 
\cellcolor[HTML]{9ac57c} 0.9579 ($\pm$0.0008) & 
\cellcolor[HTML]{bbcb75} 0.9583 ($\pm$0.0007) \\
\bottomrule
\end{tabular}
\end{adjustbox}
\caption{Black (composite) precision. Averaged performance across three models.}
\label{tab:black_composite_precision}
\end{table}

\begin{table}[H]
\centering
\begin{adjustbox}{width=\textwidth,center}
\begin{tabular}{c|ccccccc}
\toprule
\multirow{2}{*}{\textbf{Test set}} & \multicolumn{5}{c}{\textbf{Train set}} \\
\cmidrule(lr){2-8}
& 25m & 50m & 100m & 200m & 300m & 400m & 500m \\
\midrule
25m & 
\cellcolor[HTML]{bbcb75} 0.9050 ($\pm$0.0038) & 
\cellcolor[HTML]{c7cd72} 0.9084 ($\pm$0.0027) & 
\cellcolor[HTML]{d3ce70} 0.9121 ($\pm$0.0023) & 
\cellcolor[HTML]{dad06e} 0.9142 ($\pm$0.0017) & 
\cellcolor[HTML]{d7cf6f} 0.9134 ($\pm$0.0019) & 
\cellcolor[HTML]{dcd06e} 0.9149 ($\pm$0.0023) & 
\cellcolor[HTML]{e3d16c} 0.9169 ($\pm$0.0020) \\
50m & 
\cellcolor[HTML]{bdcb75} 0.9055 ($\pm$0.0066) & 
\cellcolor[HTML]{e4d16c} 0.9170 ($\pm$0.0052) & 
\cellcolor[HTML]{fdcd67} 0.9269 ($\pm$0.0036) & 
\cellcolor[HTML]{f7b76a} 0.9310 ($\pm$0.0035) & 
\cellcolor[HTML]{f6b56a} 0.9315 ($\pm$0.0033) & 
\cellcolor[HTML]{f3aa6c} 0.9336 ($\pm$0.0027) & 
\cellcolor[HTML]{f09f6e} 0.9358 ($\pm$0.0017) \\
100m & 
\cellcolor[HTML]{96c57d} 0.8941 ($\pm$0.0095) & 
\cellcolor[HTML]{d0ce70} 0.9113 ($\pm$0.0070) & 
\cellcolor[HTML]{fcc967} 0.9276 ($\pm$0.0047) & 
\cellcolor[HTML]{f7b76a} 0.9311 ($\pm$0.0056) & 
\cellcolor[HTML]{f6b36b} 0.9319 ($\pm$0.0050) & 
\cellcolor[HTML]{f0a06d} 0.9355 ($\pm$0.0041) & 
\cellcolor[HTML]{eb8e70} 0.9392 ($\pm$0.0022) \\
200m & 
\cellcolor[HTML]{57bb8a} 0.8753 ($\pm$0.0108) & 
\cellcolor[HTML]{9fc67b} 0.8967 ($\pm$0.0088) & 
\cellcolor[HTML]{e3d16c} 0.9168 ($\pm$0.0063) & 
\cellcolor[HTML]{f9d568} 0.9233 ($\pm$0.0063) & 
\cellcolor[HTML]{fad567} 0.9237 ($\pm$0.0072) & 
\cellcolor[HTML]{fbc768} 0.9280 ($\pm$0.0063) & 
\cellcolor[HTML]{f6b46a} 0.9316 ($\pm$0.0039) \\
300m & 
\cellcolor[HTML]{9fc67b} 0.8967 ($\pm$0.0078) & 
\cellcolor[HTML]{d2ce70} 0.9119 ($\pm$0.0076) & 
\cellcolor[HTML]{fdcd67} 0.9268 ($\pm$0.0060) & 
\cellcolor[HTML]{f09f6e} 0.9358 ($\pm$0.0043) & 
\cellcolor[HTML]{f1a26d} 0.9352 ($\pm$0.0057) & 
\cellcolor[HTML]{eb8c70} 0.9396 ($\pm$0.0039) & 
\cellcolor[HTML]{e67c73} 0.9425 ($\pm$0.0030) \\
400m & 
\cellcolor[HTML]{66bd87} 0.8798 ($\pm$0.0081) & 
\cellcolor[HTML]{a7c779} 0.8992 ($\pm$0.0080) & 
\cellcolor[HTML]{e2d16d} 0.9166 ($\pm$0.0066) & 
\cellcolor[HTML]{ffd666} 0.9250 ($\pm$0.0071) & 
\cellcolor[HTML]{fbc668} 0.9281 ($\pm$0.0062) & 
\cellcolor[HTML]{f4ae6b} 0.9328 ($\pm$0.0058) & 
\cellcolor[HTML]{ef9b6e} 0.9367 ($\pm$0.0040) \\
500m & 
\cellcolor[HTML]{66bd87} 0.8799 ($\pm$0.0069) & 
\cellcolor[HTML]{acc878} 0.9004 ($\pm$0.0077) & 
\cellcolor[HTML]{e9d26b} 0.9186 ($\pm$0.0072) & 
\cellcolor[HTML]{fbc768} 0.9281 ($\pm$0.0064) & 
\cellcolor[HTML]{f6b46a} 0.9317 ($\pm$0.0064) & 
\cellcolor[HTML]{f1a46d} 0.9349 ($\pm$0.0054) & 
\cellcolor[HTML]{eb8e70} 0.9391 ($\pm$0.0049) \\
\bottomrule
\end{tabular}
\end{adjustbox}
\caption{Black (composite) recall. Averaged performance across three models.}
\label{tab:black_composite_recall}
\end{table}

\begin{table}[H]
\centering
\begin{adjustbox}{width=\textwidth,center}
\begin{tabular}{c|ccccccc}
\toprule
\multirow{2}{*}{\textbf{Test set}} & \multicolumn{5}{c}{\textbf{Train set}} \\
\cmidrule(lr){2-8}
& 25m & 50m & 100m & 200m & 300m & 400m & 500m\\
\midrule
25m & 
\cellcolor[HTML]{bfcb74} 0.9324 ($\pm$0.0016) & 
\cellcolor[HTML]{c9cd72} 0.9338 ($\pm$0.0013) & 
\cellcolor[HTML]{d9d06e} 0.9362 ($\pm$0.0011) & 
\cellcolor[HTML]{e3d16c} 0.9376 ($\pm$0.0010) & 
\cellcolor[HTML]{e2d16d} 0.9374 ($\pm$0.0011) & 
\cellcolor[HTML]{e7d26c} 0.9381 ($\pm$0.0011) & 
\cellcolor[HTML]{f1d369} 0.9395 ($\pm$0.0010) \\
50m & 
\cellcolor[HTML]{b7ca76} 0.9313 ($\pm$0.0028) & 
\cellcolor[HTML]{e0d16d} 0.9371 ($\pm$0.0024) & 
\cellcolor[HTML]{fed166} 0.9420 ($\pm$0.0019) & 
\cellcolor[HTML]{f6b66a} 0.9445 ($\pm$0.0017) & 
\cellcolor[HTML]{f6b36b} 0.9448 ($\pm$0.0018) & 
\cellcolor[HTML]{f3a86c} 0.9458 ($\pm$0.0012) & 
\cellcolor[HTML]{ee996e} 0.9472 ($\pm$0.0009) \\
100m & 
\cellcolor[HTML]{94c47d} 0.9263 ($\pm$0.0044) & 
\cellcolor[HTML]{cece71} 0.9346 ($\pm$0.0033) & 
\cellcolor[HTML]{fccb67} 0.9425 ($\pm$0.0022) & 
\cellcolor[HTML]{f7b76a} 0.9444 ($\pm$0.0025) & 
\cellcolor[HTML]{f5b06b} 0.9450 ($\pm$0.0024) & 
\cellcolor[HTML]{f09f6d} 0.9466 ($\pm$0.0020) & 
\cellcolor[HTML]{e98771} 0.9488 ($\pm$0.0009) \\
200m & 
\cellcolor[HTML]{57bb8a} 0.9175 ($\pm$0.0052) & 
\cellcolor[HTML]{a1c67b} 0.9281 ($\pm$0.0041) & 
\cellcolor[HTML]{e3d16c} 0.9375 ($\pm$0.0030) & 
\cellcolor[HTML]{fcd567} 0.9412 ($\pm$0.0028) & 
\cellcolor[HTML]{fed567} 0.9413 ($\pm$0.0034) & 
\cellcolor[HTML]{fac268} 0.9434 ($\pm$0.0029) & 
\cellcolor[HTML]{f3aa6c} 0.9456 ($\pm$0.0017) \\
300m & 
\cellcolor[HTML]{9bc67c} 0.9273 ($\pm$0.0035) & 
\cellcolor[HTML]{cece71} 0.9345 ($\pm$0.0034) & 
\cellcolor[HTML]{fed066} 0.9421 ($\pm$0.0027) & 
\cellcolor[HTML]{f1a16d} 0.9464 ($\pm$0.0021) & 
\cellcolor[HTML]{f1a36d} 0.9462 ($\pm$0.0026) & 
\cellcolor[HTML]{eb8d70} 0.9483 ($\pm$0.0018) & 
\cellcolor[HTML]{e67c73} 0.9498 ($\pm$0.0013) \\
400m & 
\cellcolor[HTML]{66bd87} 0.9198 ($\pm$0.0036) & 
\cellcolor[HTML]{a8c879} 0.9292 ($\pm$0.0038) & 
\cellcolor[HTML]{e1d16d} 0.9373 ($\pm$0.0031) & 
\cellcolor[HTML]{ffd666} 0.9415 ($\pm$0.0033) & 
\cellcolor[HTML]{fac368} 0.9432 ($\pm$0.0029) & 
\cellcolor[HTML]{f3ab6c} 0.9455 ($\pm$0.0028) & 
\cellcolor[HTML]{ee966f} 0.9475 ($\pm$0.0019) \\
500m & 
\cellcolor[HTML]{65bd87} 0.9195 ($\pm$0.0034) & 
\cellcolor[HTML]{acc878} 0.9296 ($\pm$0.0037) & 
\cellcolor[HTML]{ead26b} 0.9385 ($\pm$0.0033) & 
\cellcolor[HTML]{fbc668} 0.9430 ($\pm$0.0029) & 
\cellcolor[HTML]{f5b16b} 0.9449 ($\pm$0.0030) & 
\cellcolor[HTML]{f1a26d} 0.9463 ($\pm$0.0025) & 
\cellcolor[HTML]{ea8b70} 0.9485 ($\pm$0.0021) \\
\bottomrule
\end{tabular}
\end{adjustbox}
\caption{Black (composite) F1. Averaged performance across three models.}
\label{tab:black_composite_f1}
\end{table}

The noticeable step-change in performance/behaviour between the model trained at 25m and others was also visible in the training data, and will be explored in more detail below. 

\begin{figure}[H]
    \centering
    \includegraphics[width=1\linewidth]{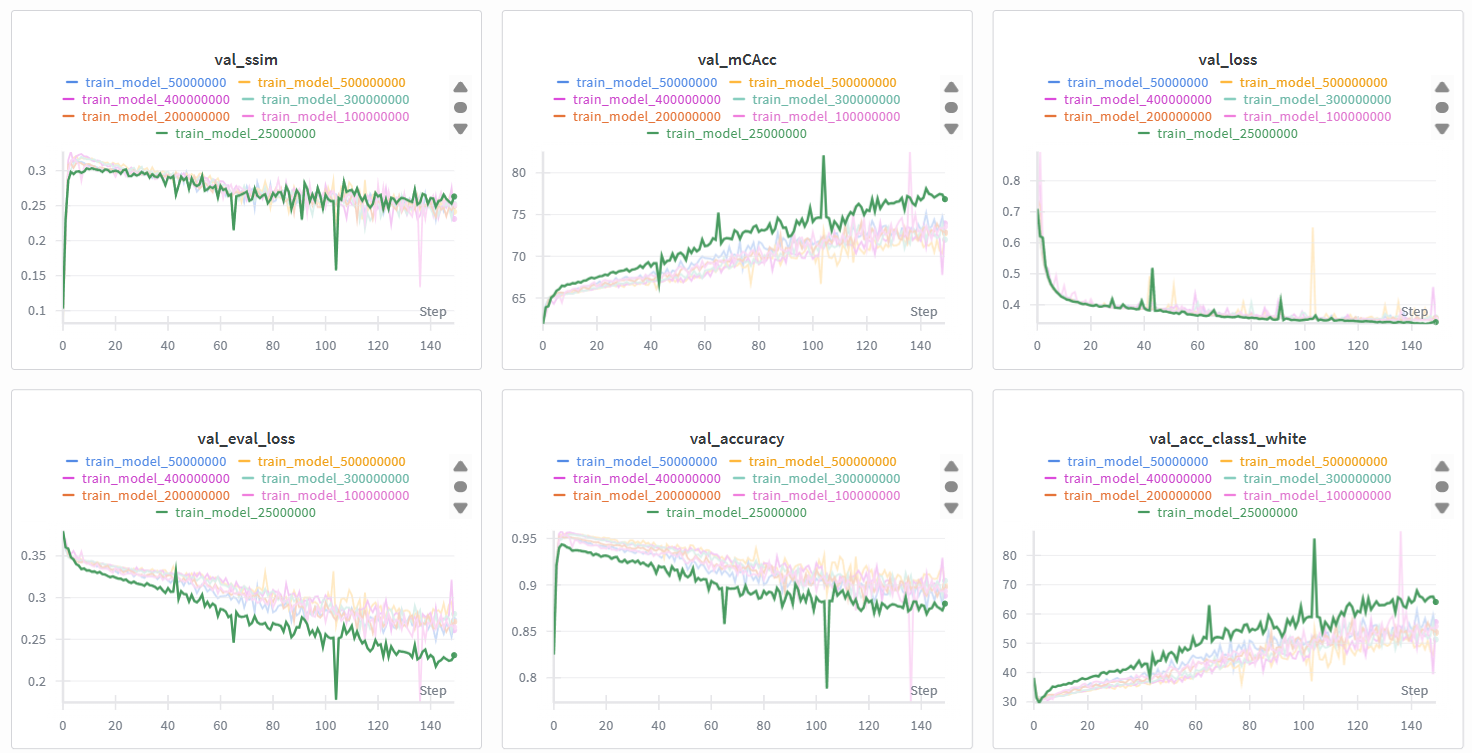}
    \includegraphics[width=1\linewidth]{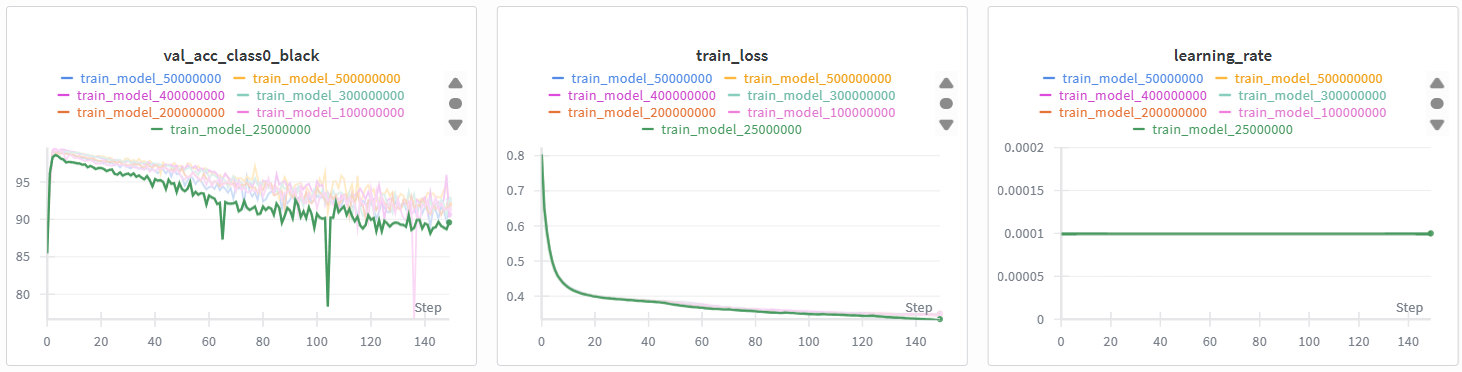}
    \caption{Training metrics for a one of the three sets of models produced. Note the observable gap between the model trained on the 25m data (dark green) and other models.}
    \label{fig:training_metrics}
\end{figure}

\section{Validation Checks}
Given the noticeable jump in performance and divergence in behaviours between the model trained at 25m and those trained at higher bands, it was necessary to check whether the effects described above were not simply a reflection of the higher density of prime pixels closer to zero. To study the contribution of pixel class imbalance to the observed effects, we performed two additional experiments:
\begin{itemize}
    \item \textbf{Naïve ratio-aligned baseline}: We compared the models scores against those expected from a baseline predictor with knowledge of the approximate black:white ratio in a given region, but which assigned pixel classes randomly subject to this proportion. In other words, we calculated the expected results if a deterministic algorithm were to assign roughly 5\% of all pixels in a block from the 500m region as white, but selected the specific pixels entirely at random.
    \item \textbf{Top-k binarization}: We evaluated the trained models under a constrained decoding scheme, where the top 6\% of predicted logits (by value) were forcibly assigned to the prime (white) class, irrespective of their absolute probability.
\end{itemize}

Together, these checks allowed us to distinguish genuine structural learning effects from those explainable purely by skew in the class distribution.

\subsection{Naïve Ratio-Aligned Baseline}
For blocks in the 25m range, the proportion of white pixels is over 6\%, it hovers around 6\% at the 50m range, and majority of blocks from other ranges have between 5 and 6\% white pixels, a distribution that can be roughly graphed using the prime number theorem.

\begin{figure}[H]
    \centering
    \includegraphics[width=1\linewidth]{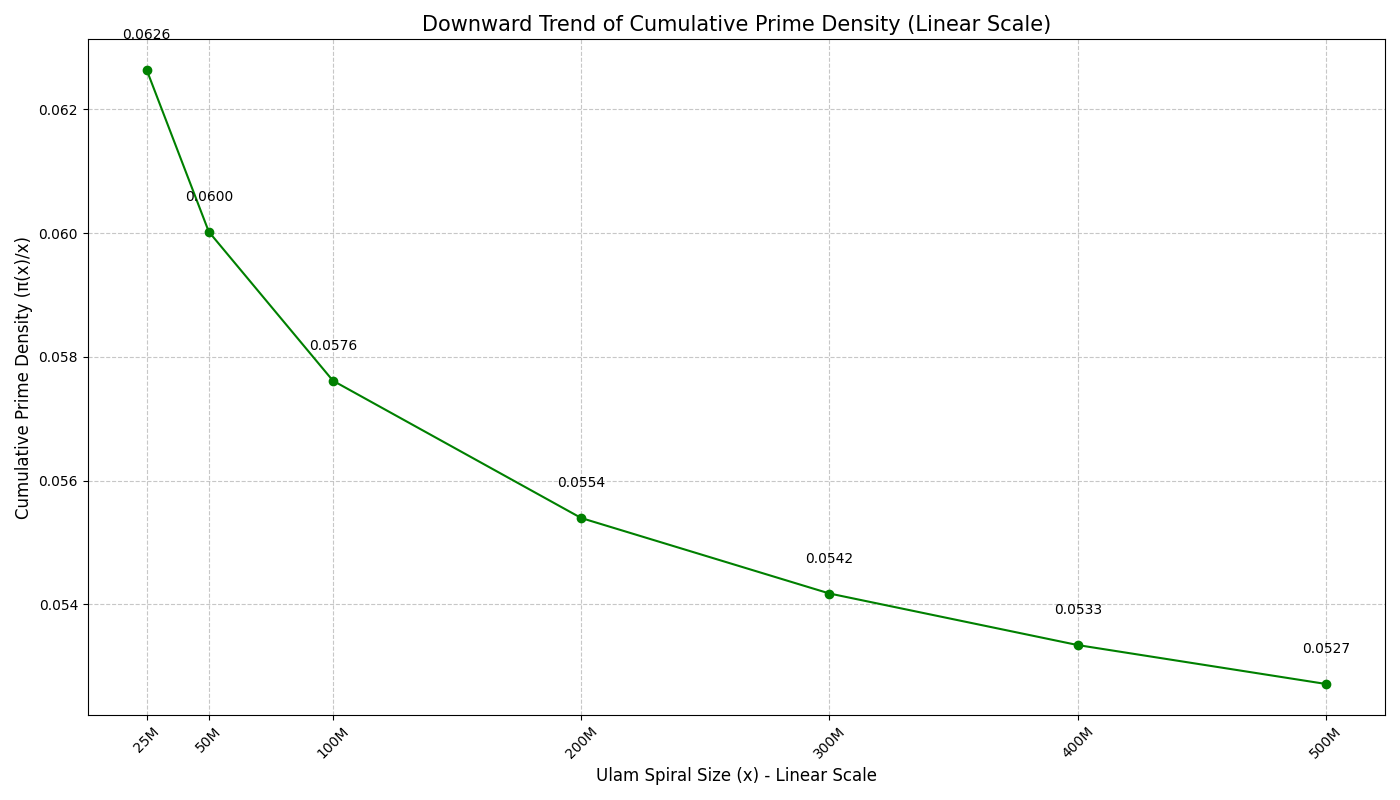}
    \caption{Density of primes lower than x calculated per the prime number theorem.}
    \label{fig:prime_density}
\end{figure}

Assuming the model learnt only the likely proportion of black and white pixels in a given image without learning any patterns, the expected result distribution would be as follows:

\begin{table}[H]
\centering
\begin{adjustbox}{width=\textwidth,center}
\begin{tabular}{|l|c|c|c|c|}
\hline & 
\textbf{\makecell{Prime prevalence = 5.27\%, \\ Assigned prime = 5.27\%}} & 
\textbf{\makecell{Prime prevalence = 5.27\%, \\ Assigned prime = 6.26\%}} & 
\textbf{\makecell{Prime Prevalence = 6.26\%, \\ assigned prime = 5.27\%}} & 
\textbf{\makecell{Prime Prevalence = 6.26\%, \\ assigned prime = 6.26\%}} \\
\hline
Overall accuracy/micro F1 & 
0.900155 & 
0.891298 & 
0.891298 & 
0.882638 \\
\hline
Macro F1 & 
0.5 & 
0.499395 & 
0.499774 & 
0.5 \\
\hline
Prime precision & 
0.0527 & 
0.0527 & 
0.0626 & 
0.0626 \\
\hline
Prime recall & 
0.0527 & 
0.0626 & 
0.0527 & 
0.0626 \\
\hline
Prime F1 & 
0.0527 & 
0.057225 & 
0.057225 & 
0.0626 \\
\hline
Composite precision & 
0.9473 & 
0.9374 &
0.9473 & 
0.9374 \\
\hline
Composite F1 & 
0.9473 & 
0.942324 & 
0.942324 & 
0.9374 \\
\hline
\end{tabular}
\end{adjustbox}
\caption{Expected outcomes in low/high spiral regions assuming zero knowledge of structure but awareness of average white-to-black ratios.}
\label{tab:expected_outcomes}
\end{table}

Firstly, and most obviously, it is worth noting that while a model that simply guessed based on white:black ratios in a given region without learning patterns would achieve similar black and overall scores to the model we describe above, it would score significantly lower on white metrics, suggesting that our model shows a degree of pattern-awareness.

However, it is also possible to use the estimates above to evaluate the contribution of prime density awareness (as opposed to more detailed pattern-learning) to the overall model performance.  As mentioned above, a model that simply assigns 94-95\% of all pixels black could achieve a high accuracy/micro F1 score - hence our focus on decomposing precision, recall, class F1 and confusion metrics in the present paper. By calculating the results of naïve pixel assignment at different assumed/actual prime densities, we can establish a rough idea of the relative contribution of prime density and prime patterns to the overall result achieved by the model (with the proviso that the two factors are not fully independent variables - higher prime density at low numbers is partly a product of prime patterns).

It is likewise possible to chart the rate of change in the prime density/accuracy of naïve ratio-based prediction (these two curves being identical when normalised to a single index as here) and thus compare it to the decline in prime F1 scores of each model trained.

\begin{figure}[H]
    \centering
    \includegraphics[width=1\linewidth]{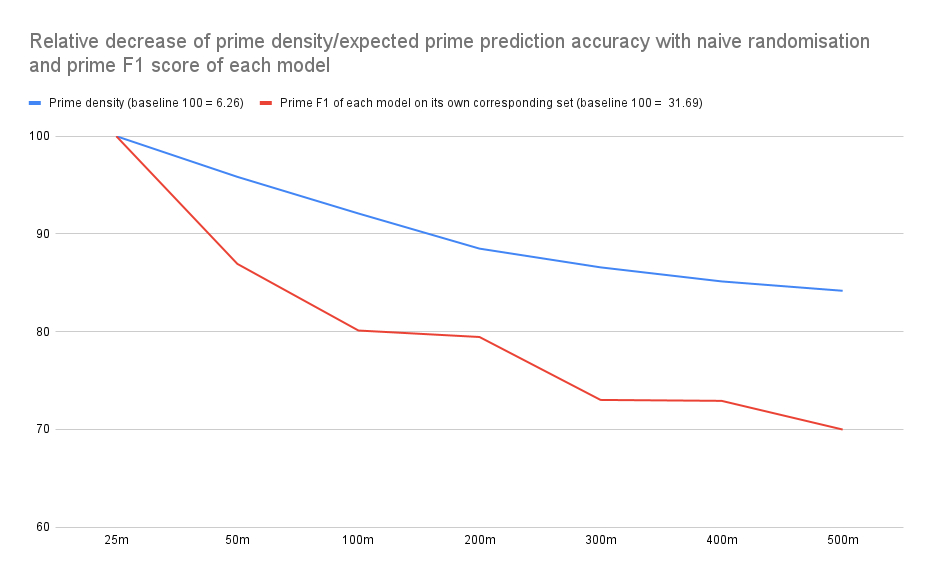}
    \caption{Comparative drop in prime density/accuracy of naïve ratio-based prediction and models' F1 scores when run on the test set corresponding to their local training set.}
    \label{fig:f1_drop}
\end{figure}

The models' ability to predict primes declines faster at higher numbers than would be expected if they were relying entirely upon white/black ratios to make their predictions. Consequently we can conclude that the model prime prediction performance bonus is associated with pattern learning, but also that the patterns it is learning dissipate faster than the total proportion of primes drops.

\subsection{Top-k Binarisation}
To evaluate the precise proportion of the model’s scores attributable to learning black-to-white ratios as opposed to understanding pixel structure, we proceeded to train a second model using quota/top-k binarisation. This selected the top 5\% or 6\% of logits/probabilities per image as white, regardless of their absolute value, helping to isolate the the possible confound created by the white-to-black imbalance.The example results given here show the outcome if the top 6\% of pixels are set to white.

\begin{table}[H]
\centering
\begin{adjustbox}{width=\textwidth,center}
\begin{tabular}{c|ccccccc}
\toprule
\multirow{2}{*}{\textbf{Test set}} & \multicolumn{5}{c}{\textbf{Train set}} \\
\cmidrule(lr){2-8}
& 25m & 50m & 100m & 200m & 300m & 400m & 500m \\
\midrule
25m & 
\cellcolor[HTML]{aac879} 0.917 ($\pm$0.001) & 
\cellcolor[HTML]{a0c67b} 0.917 ($\pm$0.001) & 
\cellcolor[HTML]{e6d26c} 0.918 ($\pm$0.001) & 
\cellcolor[HTML]{fcca67} 0.919 ($\pm$0.001) & 
\cellcolor[HTML]{f5af6b} 0.919 ($\pm$0.000) & 
\cellcolor[HTML]{f2a76c} 0.920 ($\pm$0.001) & 
\cellcolor[HTML]{e67c73} 0.920 ($\pm$0.000) \\
50m & 
\cellcolor[HTML]{65bd87} 0.915 ($\pm$0.001) & 
\cellcolor[HTML]{9fc67b} 0.917 ($\pm$0.001) & 
\cellcolor[HTML]{e3d16c} 0.918 ($\pm$0.001) & 
\cellcolor[HTML]{ffd466} 0.919 ($\pm$0.001) & 
\cellcolor[HTML]{f8bc69} 0.919 ($\pm$0.001) & 
\cellcolor[HTML]{f5b06b} 0.919 ($\pm$0.001) & 
\cellcolor[HTML]{e88172} 0.920 ($\pm$0.000) \\
100m & 
\cellcolor[HTML]{61bc88} 0.915 ($\pm$0.001) & 
\cellcolor[HTML]{9cc67c} 0.917 ($\pm$0.001) & 
\cellcolor[HTML]{e5d16c} 0.918 ($\pm$0.001) & 
\cellcolor[HTML]{fdce67} 0.919 ($\pm$0.001) & 
\cellcolor[HTML]{f9bf69} 0.919 ($\pm$0.001) & 
\cellcolor[HTML]{f5af6b} 0.919 ($\pm$0.001) & 
\cellcolor[HTML]{e78072} 0.920 ($\pm$0.000)\\
200m & 
\cellcolor[HTML]{57bb8a} 0.915 ($\pm$0.001) & 
\cellcolor[HTML]{90c47e} 0.917 ($\pm$0.001) & 
\cellcolor[HTML]{d8cf6f} 0.918 ($\pm$0.001) & 
\cellcolor[HTML]{ffd666} 0.919 ($\pm$0.001) & 
\cellcolor[HTML]{fac268} 0.919 ($\pm$0.001) & 
\cellcolor[HTML]{f6b36b} 0.919 ($\pm$0.001) & 
\cellcolor[HTML]{ea8971} 0.920 ($\pm$0.001) \\
300m & 
\cellcolor[HTML]{5bbb8a} 0.915 ($\pm$0.001) & 
\cellcolor[HTML]{90c47e} 0.916 ($\pm$0.001) & 
\cellcolor[HTML]{d8cf6f} 0.918 ($\pm$0.001) & 
\cellcolor[HTML]{f9d568} 0.919 ($\pm$0.001) & 
\cellcolor[HTML]{fcca67} 0.919 ($\pm$0.001) & 
\cellcolor[HTML]{f6b56a} 0.919 ($\pm$0.001) & 
\cellcolor[HTML]{eb8c70} 0.920 ($\pm$0.000) \\
400m & 
\cellcolor[HTML]{57bb8a} 0.915 ($\pm$0.001) & 
\cellcolor[HTML]{8fc47e} 0.916 ($\pm$0.001) & 
\cellcolor[HTML]{d4cf70} 0.918 ($\pm$0.001) & 
\cellcolor[HTML]{f4d469} 0.919 ($\pm$0.001) & 
\cellcolor[HTML]{fdcf67} 0.919 ($\pm$0.001) & 
\cellcolor[HTML]{fac368} 0.919 ($\pm$0.001) & 
\cellcolor[HTML]{ec9170} 0.920 ($\pm$0.000) \\
500m & 
\cellcolor[HTML]{5bbb8a} 0.915 ($\pm$0.001) & 
\cellcolor[HTML]{92c47e} 0.916 ($\pm$0.001) & 
\cellcolor[HTML]{d5cf6f} 0.918 ($\pm$0.001) & 
\cellcolor[HTML]{fad567} 0.919 ($\pm$0.001) & 
\cellcolor[HTML]{fcc967} 0.919 ($\pm$0.001) & 
\cellcolor[HTML]{fac368} 0.919 ($\pm$0.001) & 
\cellcolor[HTML]{eb8b70} 0.920 ($\pm$0.000) \\
\bottomrule
\end{tabular}
\end{adjustbox}
\caption{Overall accuracy/micro F1. Averaged performance across three models.}
\label{tab:binarisation_overall_accuracy_micro_f1}
\end{table}

\begin{table}[H]
\centering
\begin{adjustbox}{width=\textwidth,center}
\begin{tabular}{c|ccccccc}
\toprule
\multirow{2}{*}{\textbf{Test set}} & \multicolumn{5}{c}{\textbf{Train set}} \\
\cmidrule(lr){2-8}
& 25m & 50m & 100m & 200m & 300m & 400m & 500m \\
\midrule
25m & 
\cellcolor[HTML]{e67c73} 0.583 ($\pm$0.002) & 
\cellcolor[HTML]{fac268} 0.575 ($\pm$0.001) & 
\cellcolor[HTML]{fbc568} 0.574 ($\pm$0.002) & 
\cellcolor[HTML]{fccb67} 0.574 ($\pm$0.001) & 
\cellcolor[HTML]{fdce67} 0.573 ($\pm$0.001) & 
\cellcolor[HTML]{fed366} 0.573 ($\pm$0.001) & 
\cellcolor[HTML]{ffd666} 0.572 ($\pm$0.001) \\
50m & 
\cellcolor[HTML]{fac468} 0.575 ($\pm$0.001) & 
\cellcolor[HTML]{fbc668} 0.574 ($\pm$0.001) & 
\cellcolor[HTML]{fcc967} 0.574 ($\pm$0.001) & 
\cellcolor[HTML]{ffd566} 0.572 ($\pm$0.001) & 
\cellcolor[HTML]{f2d369} 0.572 ($\pm$0.001) & 
\cellcolor[HTML]{e9d26b} 0.572 ($\pm$0.001) & 
\cellcolor[HTML]{cacd72} 0.572 ($\pm$0.001) \\
100m & 
\cellcolor[HTML]{fbc868} 0.574 ($\pm$0.001) & 
\cellcolor[HTML]{fcca67} 0.574 ($\pm$0.001) & 
\cellcolor[HTML]{fcca67} 0.574 ($\pm$0.001) & 
\cellcolor[HTML]{fed066} 0.573 ($\pm$0.001) & 
\cellcolor[HTML]{f6d468} 0.572 ($\pm$0.001) & 
\cellcolor[HTML]{f6d468} 0.572 ($\pm$0.001) & 
\cellcolor[HTML]{e0d16d} 0.572 ($\pm$0.001) \\
200m & 
\cellcolor[HTML]{fed266} 0.573 ($\pm$0.001) &
\cellcolor[HTML]{fed166} 0.573 ($\pm$0.001) & 
\cellcolor[HTML]{ffd366} 0.573 ($\pm$0.001) & 
\cellcolor[HTML]{ffd666} 0.572 ($\pm$0.001) & 
\cellcolor[HTML]{cdce71} 0.572 ($\pm$0.001) & 
\cellcolor[HTML]{d9cf6f} 0.572 ($\pm$0.001) & 
\cellcolor[HTML]{a7c779} 0.571 ($\pm$0.001) \\
300m & 
\cellcolor[HTML]{fdce67} 0.573 ($\pm$0.001) & 
\cellcolor[HTML]{ffd666} 0.572 ($\pm$0.001) & 
\cellcolor[HTML]{ffd666} 0.572 ($\pm$0.001) & 
\cellcolor[HTML]{d0ce70} 0.572 ($\pm$0.001) & 
\cellcolor[HTML]{97c57d} 0.571 ($\pm$0.001) & 
\cellcolor[HTML]{bccb75} 0.571 ($\pm$0.001) & 
\cellcolor[HTML]{7dc182} 0.571 ($\pm$0.001) \\
400m & 
\cellcolor[HTML]{fed166} 0.573 ($\pm$0.001) & 
\cellcolor[HTML]{fbd567} 0.572 ($\pm$0.001) & 
\cellcolor[HTML]{d9d06e} 0.572 ($\pm$0.001) & 
\cellcolor[HTML]{8bc37f} 0.571 ($\pm$0.001) & 
\cellcolor[HTML]{5fbc89} 0.570 ($\pm$0.001) & 
\cellcolor[HTML]{79c083} 0.571 ($\pm$0.001) & 
\cellcolor[HTML]{57bb8a} 0.570 ($\pm$0.001) \\
500m & 
\cellcolor[HTML]{fed066} 0.573 ($\pm$0.001) & 
\cellcolor[HTML]{ffd566} 0.572 ($\pm$0.001) & 
\cellcolor[HTML]{d2ce70} 0.572 ($\pm$0.001) & 
\cellcolor[HTML]{c2cc74} 0.571 ($\pm$0.001) & 
\cellcolor[HTML]{89c380} 0.571 ($\pm$0.001) & 
\cellcolor[HTML]{64bd88} 0.570 ($\pm$0.001) & 
\cellcolor[HTML]{78c083} 0.571 ($\pm$0.001) \\
\bottomrule
\end{tabular}
\end{adjustbox}
\caption{Macro F1. Averaged performance across three models.}
\label{tab:binarisation_macro_f1}
\end{table}

\begin{table}[H]
\centering
\begin{adjustbox}{width=\textwidth,center}
\begin{tabular}{c|ccccccc}
\toprule
\multirow{2}{*}{\textbf{Test set}} & \multicolumn{5}{c}{\textbf{Train set}} \\
\cmidrule(lr){2-8}
& 25m & 50m & 100m & 200m & 300m & 400m & 500m \\
\midrule
25m & 
\cellcolor[HTML]{e67c73} 0.211 ($\pm$0.004) & 
\cellcolor[HTML]{f9bf69} 0.193 ($\pm$0.002) & 
\cellcolor[HTML]{fbc668} 0.192 ($\pm$0.003) & 
\cellcolor[HTML]{fdcc67} 0.190 ($\pm$0.003) & 
\cellcolor[HTML]{fed066} 0.189 ($\pm$0.003) & 
\cellcolor[HTML]{ffd566} 0.188 ($\pm$0.003) & 
\cellcolor[HTML]{e6d26c} 0.187 ($\pm$0.002) \\
50m & 
\cellcolor[HTML]{f9be69} 0.194 ($\pm$0.002) &
\cellcolor[HTML]{fac368} 0.192 ($\pm$0.002) & 
\cellcolor[HTML]{fcc967} 0.191 ($\pm$0.002) & 
\cellcolor[HTML]{ffd666} 0.188 ($\pm$0.003) & 
\cellcolor[HTML]{e6d26c} 0.187 ($\pm$0.002) & 
\cellcolor[HTML]{dbd06e} 0.186 ($\pm$0.002) & 
\cellcolor[HTML]{b3c977} 0.185 ($\pm$0.002) \\
100m & 
\cellcolor[HTML]{fac169} 0.193 ($\pm$0.002) & 
\cellcolor[HTML]{fbc668} 0.192 ($\pm$0.002) & 
\cellcolor[HTML]{fcca67} 0.190 ($\pm$0.003) & 
\cellcolor[HTML]{fed166} 0.189 ($\pm$0.002) & 
\cellcolor[HTML]{ead26b} 0.187 ($\pm$0.002) & 
\cellcolor[HTML]{e6d16c} 0.187 ($\pm$0.002) & 
\cellcolor[HTML]{c5cc73} 0.186 ($\pm$0.002) \\
200m & 
\cellcolor[HTML]{fccb67} 0.190 ($\pm$0.003) & 
\cellcolor[HTML]{fdcd67} 0.190 ($\pm$0.002) & 
\cellcolor[HTML]{fed266} 0.188 ($\pm$0.003) & 
\cellcolor[HTML]{ffd666} 0.187 ($\pm$0.003) & 
\cellcolor[HTML]{c9cd72} 0.186 ($\pm$0.002) & 
\cellcolor[HTML]{cece71} 0.186 ($\pm$0.003) & 
\cellcolor[HTML]{98c57d} 0.184 ($\pm$0.003) \\
300m & 
\cellcolor[HTML]{fbc768} 0.191 ($\pm$0.002) & 
\cellcolor[HTML]{fed166} 0.189 ($\pm$0.002) & 
\cellcolor[HTML]{ffd566} 0.188 ($\pm$0.002) & 
\cellcolor[HTML]{d4cf70} 0.186 ($\pm$0.002) & 
\cellcolor[HTML]{9ec67b} 0.185 ($\pm$0.002) & 
\cellcolor[HTML]{b7ca76} 0.185 ($\pm$0.002) & 
\cellcolor[HTML]{75bf84} 0.183 ($\pm$0.002) \\
400m & 
\cellcolor[HTML]{fcca67} 0.191 ($\pm$0.002) & 
\cellcolor[HTML]{fed266} 0.188 ($\pm$0.002) & 
\cellcolor[HTML]{ecd36a} 0.187 ($\pm$0.003) & 
\cellcolor[HTML]{9cc67c} 0.185 ($\pm$0.002) & 
\cellcolor[HTML]{70bf85} 0.183 ($\pm$0.003) & 
\cellcolor[HTML]{83c281} 0.184 ($\pm$0.002) & 
\cellcolor[HTML]{57bb8a} 0.183 ($\pm$0.002) \\
500m & 
\cellcolor[HTML]{fcc967} 0.191 ($\pm$0.002) & 
\cellcolor[HTML]{fed066} 0.189 ($\pm$0.002) & 
\cellcolor[HTML]{e5d16c} 0.187 ($\pm$0.002) & 
\cellcolor[HTML]{c7cd72} 0.186 ($\pm$0.002) & 
\cellcolor[HTML]{92c47e} 0.184 ($\pm$0.002) & 
\cellcolor[HTML]{71bf85} 0.183 ($\pm$0.002) & 
\cellcolor[HTML]{71bf85} 0.183 ($\pm$0.003) \\
\bottomrule
\end{tabular}
\end{adjustbox}
\caption{White F1. Averaged performance across three models.}
\label{tab:binarisation_white_f1}
\end{table}

\begin{table}[H]
\centering
\begin{adjustbox}{width=\textwidth,center}
\begin{tabular}{c|ccccccc}
\toprule
\multirow{2}{*}{\textbf{Test set}} & \multicolumn{5}{c}{\textbf{Train set}} \\
\cmidrule(lr){2-8}
& 25m & 50m & 100m & 200m & 300m & 400m & 500m \\
\midrule
25m & 
\cellcolor[HTML]{a5c77a} 0.956 ($\pm$0.000) & 
\cellcolor[HTML]{9fc67b} 0.956 ($\pm$0.000) & 
\cellcolor[HTML]{e6d26c} 0.957 ($\pm$0.000) & 
\cellcolor[HTML]{fcca67} 0.957 ($\pm$0.000) & 
\cellcolor[HTML]{f5b06b} 0.958 ($\pm$0.000) & 
\cellcolor[HTML]{f2a86c} 0.958 ($\pm$0.000) & 
\cellcolor[HTML]{e67c73} 0.958 ($\pm$0.000) \\
50m & 
\cellcolor[HTML]{64bd88} 0.955 ($\pm$0.000) & 
\cellcolor[HTML]{9fc67b} 0.956 ($\pm$0.000) & 
\cellcolor[HTML]{e2d16d} 0.957 ($\pm$0.000) & 
\cellcolor[HTML]{ffd466} 0.957 ($\pm$0.000) & 
\cellcolor[HTML]{f8bb69} 0.958 ($\pm$0.000) & 
\cellcolor[HTML]{f5b06b} 0.958 ($\pm$0.000) & 
\cellcolor[HTML]{e88072} 0.958 ($\pm$0.000) \\
100m & 
\cellcolor[HTML]{60bc88} 0.955 ($\pm$0.000) & 
\cellcolor[HTML]{9bc67c} 0.956 ($\pm$0.000) & 
\cellcolor[HTML]{e2d16d} 0.957 ($\pm$0.000) & 
\cellcolor[HTML]{fdcf67} 0.957 ($\pm$0.000) & 
\cellcolor[HTML]{f9bf69} 0.957 ($\pm$0.000) & 
\cellcolor[HTML]{f5af6b} 0.958 ($\pm$0.000) & 
\cellcolor[HTML]{e78072} 0.958 ($\pm$0.000) \\
200m & 
\cellcolor[HTML]{57bb8a} 0.955 ($\pm$0.000) & 
\cellcolor[HTML]{96c57d} 0.956 ($\pm$0.000) & 
\cellcolor[HTML]{dbd06e} 0.957 ($\pm$0.000) & 
\cellcolor[HTML]{ffd666} 0.957 ($\pm$0.000) & 
\cellcolor[HTML]{fac268} 0.957 ($\pm$0.000) & 
\cellcolor[HTML]{f5b26b} 0.958 ($\pm$0.000) & 
\cellcolor[HTML]{ea8871} 0.958 ($\pm$0.000) \\
300m & 
\cellcolor[HTML]{5bbb8a} 0.955 ($\pm$0.000) & 
\cellcolor[HTML]{91c47e} 0.956 ($\pm$0.000) & 
\cellcolor[HTML]{d8cf6f} 0.957 ($\pm$0.000) & 
\cellcolor[HTML]{fad568} 0.957 ($\pm$0.000) & 
\cellcolor[HTML]{fcc967} 0.957 ($\pm$0.000) & 
\cellcolor[HTML]{f6b56a} 0.958 ($\pm$0.000) & 
\cellcolor[HTML]{ea8b70} 0.958 ($\pm$0.000) \\
400m & 
\cellcolor[HTML]{57bb8a} 0.955 ($\pm$0.000) & 
\cellcolor[HTML]{90c47e} 0.956 ($\pm$0.000) & 
\cellcolor[HTML]{d4cf70} 0.957 ($\pm$0.000) & 
\cellcolor[HTML]{f5d469} 0.957 ($\pm$0.000) & 
\cellcolor[HTML]{fdce67} 0.957 ($\pm$0.000) & 
\cellcolor[HTML]{fac169} 0.957 ($\pm$0.000) & 
\cellcolor[HTML]{ec8f70} 0.958 ($\pm$0.000) \\
500m & 
\cellcolor[HTML]{5abb8a} 0.955 ($\pm$0.000) & 
\cellcolor[HTML]{92c47e} 0.956 ($\pm$0.000) & 
\cellcolor[HTML]{d6cf6f} 0.957 ($\pm$0.000) & 
\cellcolor[HTML]{fad567} 0.957 ($\pm$0.000) & 
\cellcolor[HTML]{fcc868} 0.957 ($\pm$0.000) & 
\cellcolor[HTML]{fac268} 0.957 ($\pm$0.000) & 
\cellcolor[HTML]{ea8a70} 0.958 ($\pm$0.000) \\
\bottomrule
\end{tabular}
\end{adjustbox}
\caption{Black F1. Averaged performance across three models.}
\label{tab:binarisation_black_f1}
\end{table}

To verify this analysis we re-ran the process using a quota/top-k binarization approach, instead of the thresholding applied to the original model. This selected the top 5\% or 6\% of logits/probabilities per image as white during both training and testing, regardless of their absolute value, helping to isolate the possible confound created by the white-to-black imbalance. As can be seen below, forcing the model to assign 6\% of pixels white results in similar overall trends in the data (albeit in a smoothed form), implying that distribution laws are learnt during the course of pattern learning.

\section{Analysis}
The observation that a model trained on the $n \approx 5 \times 10^{8}$ region achieves superior overall performance, even when tested on much smaller ranges, suggests that the large-n Ulam spiral embodies more fundamental structural regularities than its low-n counterpart. In particular, the chaotic irregularities of the small primes appear to “average out” with increasing n, giving rise to a statistical geometry that is more stable, predictable, and amenable to learning. In this sense, the spiral exhibits an asymptotic regime in which noise is suppressed and global structure dominates.

Potentially even more interesting than this, however, is the fact that this overall score is a product of higher performance on prime prediction and relatively low numbers combined with higher performance on composite prediction at higher numbers. At lower numbers - where known patterns such as the modular fingerprints of the small primes dominate - the model places more (but not complete) reliance upon these to assign primality, hence the comparatively high white F1 at low numbers. At higher numbers, however, the model begins to favour an elimination approach, identifying likely composite numbers and assigning everything else prime by default, reflected in the higher black F1 at high numbers. This suggests that prime patterns may be dissipating even faster than the raw accuracy scores imply.

Viewed through the lens of the Cramér model, the network may be exploiting deviations from the idealized “primes-as-random” process. Since the Cramér model improves as $n \to \infty$, the network’s enhanced learnability at larger scales can be interpreted as the detection of small but systematic biases in this pseudo-randomness—biases that become more visible once local fingerprints and spikes diminish.

\section{Future Research}
If the Ulam spiral does indeed admit an asymptotic geometry, then the misclassifications of the model, its false positives, may be more mathematically informative than its successes. Composites repeatedly misidentified as primes could signal locations where the spiral’s geometry aligns with prime-supporting structures. These may include quadratic polynomial diagonals with unusually high prime incidence, or classes of “near-primes” such as semiprimes and products of small factors, which statistically shadow prime distributions. A systematic analysis of these errors could yield insights into how prime density is modulated along specific algebraic loci.

Further work should also investigate whether the representations learned by models at different scales (e.g., 25M vs. 300M) correspond to distinct number-theoretic phenomena: the former potentially encoding local irregularities such as small prime gaps, the latter encoding asymptotic density laws. Comparing these internal representations may help isolate the features that distinguish transient fluctuations from enduring structural regularities.

More broadly, this line of inquiry raises the possibility that deep learning models, when trained on sufficiently large-scale encodings such as the Ulam spiral, could act as empirical probes for asymptotic number theory. Just as heuristic models like Cramér’s provide statistical predictions of prime distribution, neural networks may reveal emergent regularities not easily accessible to analytic techniques, particularly in transitional regions between local chaos and global order. In future papers we propose to investigate these techniques for the purpose of identifying areas of strong and weak primes and semiprimes for cryptographic purposes.

\section{Annex 1: Model}
\subsection{Model Architecture}

We employ a two-dimensional U-Net architecture (\texttt{segmentation\_models\_pytorch}) with a ResNet-34 encoder pretrained on ImageNet. The model takes single-channel (grayscale) images of size $256 \times 256$ as input and outputs a single-channel probability map via a sigmoid activation. This formulates the task as binary pixel-wise prediction, where the network learns to reconstruct the full binary image from sparsely observed pixels.

\begin{itemize}
    \item Encoder: ResNet-34, pretrained on ImageNet
    \item Decoder: standard U-Net upsampling with skip connections
    \item Input channels: 1 (grayscale)
    \item Output channels: 1 (binary probability map)
    \item Activation: Sigmoid
\end{itemize}

\subsection{Data and Preprocessing}

Training data are drawn from grayscale image patches, resized to $256\times256$. Each image is normalized to $[0,1]$. For every training iteration, a Bernoulli mask is applied such that approximately 30\% of pixels remain visible while 70\% are masked. The masked image serves as input, and the original full image is used as the reconstruction target.

\subsection{Self-Supervised Objective}

The loss combines a class-balanced soft Mean Class Accuracy (Soft-MCA) with binary cross-entropy (BCE):

\[
\mathcal{L} = \alpha\, \mathcal{L}_{\text{Soft-MCA}} + \beta\, \text{BCE}, \quad \alpha=1.0,\, \beta=0.5.
\]

Soft-MCA computes class-balanced accuracy without thresholding, mitigating class imbalance. BCE enforces pixel-wise fidelity. For evaluation, we also compute hard-thresholded (0.5) Mean Class Accuracy.

\subsection{Optimization and Training}

The model is optimized with Adam at a learning rate of $10^{-4}$ and trained with batch size 4 for 150 epochs. A ReduceLROnPlateau scheduler halves the learning rate after 3 epochs of non-improvement in validation loss, with the final model being is selected by best mean accuracy score. Random seeds (42) are set for reproducibility across Python, NumPy, and PyTorch.

\subsection{Evaluation Metrics}

We report several metrics each epoch:

\begin{itemize}
    \item Pixel accuracy after thresholding at 0.5
    \item Mean Class Accuracy (hard MCA) after thresholding
    \item Structural Similarity Index Measure (SSIM) between predictions and ground truth
    \item Validation loss components (Soft-MCA, BCE)
\end{itemize}

Qualitative monitoring is performed by saving example outputs (target, masked input, prediction) each epoch.

\subsection{Conceptual Function}

The network is designed to infer global structure from sparse evidence: given only 30\% of randomly revealed pixels, it predicts the full binary image. This encourages the encoder--decoder to capture long-range spatial dependencies and internalize statistical regularities in the data. The class-balanced loss ensures robust performance across both foreground and background classes.

\section*{Acknowledgments}

\noindent This research was conducted with the assistance of various AI models: GPT-4o, 3o and 5, Claude 4 Sonnet, Gemini 2.5 Flash and Grok.

\section*{Replication}

Github: \url{https://github.com/CloudStriker/ulam-spiral-inpainting}

\end{document}